\documentclass[letterpaper, 10 pt, conference]{ieeeconf}  

\IEEEoverridecommandlockouts                              
\usepackage{cite}
\usepackage[pdftex]{graphicx}
\usepackage[cmex10]{amsmath}
\usepackage{amssymb}
\usepackage{import}
\usepackage{bm}
\usepackage{color}
\usepackage{url}
\usepackage{float}
\usepackage{epstopdf}
\usepackage{stmaryrd}
\usepackage{siunitx}
\usepackage{outlines}
\usepackage{boldline}
\usepackage{verbatim}
\usepackage[flushleft]{threeparttable}

\usepackage[utf8]{inputenc}

\usepackage{mathtools}
\usepackage{color,soul}
\usepackage{ntheorem}

\usepackage{algorithm}
\usepackage{algorithmicx}
\usepackage{algpseudocode}
\usepackage{etoolbox}\AtBeginEnvironment{algorithmic}{\small}

\usepackage{balance}
\usepackage{subcaption}

\usepackage{multirow}

\newcolumntype{?}[1]{!{\vrule width #1}}
\theoremstyle{remark}

\newcommand{\mnoshow}[1]{}

\newcommand{\figref}[1]{Fig.~\ref{#1}}

\newcolumntype{?}[1]{!{\vrule width #1}}
\theoremstyle{remark}

\newcommand*{\horzbar}{\rule[.5ex]{2.5ex}{0.5pt}}

\title{\LARGE \bf
A Distributed Multi-Vehicle Coordination Algorithm for Navigation in Tight Environments
}

\author{Roya Firoozi, 
Laura Ferranti, Xiaojing Zhang, Sebastian Nejadnik, Francesco Borrelli
\thanks{The authors are with the Department of Mechanical Engineering, at the University of California, Berkeley and Delft University of Technology. 
\newline  {\tt \{royafiroozi, xiaojing.zhang, snejadnik   ,fborrelli \}@berkeley.edu}, {\tt l.ferranti@tudelft.nl}
\newline This work has been partially supported by the Dutch Science Foundation
NWO-TTW within the Veni project HARMONIA (nr. 18165).}%
}

\begin{document}

\maketitle
\thispagestyle{empty}
\pagestyle{empty}
 
\begin{abstract}
This work presents a distributed method for multi-vehicle coordination based on nonlinear model predictive control (NMPC) and dual decomposition. Our approach allows the vehicles to coordinate in tight spaces (e.g., busy highway lanes or parking lots) by using a polytopic description of each vehicle's shape and formulating collision avoidance as a dual optimization problem. Our method accommodates heterogeneous teams of vehicles (i.e., vehicles with different polytopic shapes and dynamic models can be part of the same team). Our method allows the vehicles to share their intentions in a distributed fashion without relying on a central coordinator and efficiently provides collision-free trajectories for the vehicles. In addition, our method decouples the individual-vehicles' trajectory optimization from their collision-avoidance objectives enhancing the scalability of the method and allowing one to exploit parallel hardware architectures. All these features are particularly important for vehicular applications, where the systems operate at high-frequency rates in dynamic environments.
To validate our method, we apply it in a vehicular application, that is, the autonomous lane-merging of a team of connected vehicles to form a platoon. We compare our design with the centralized NMPC design to show the computational benefits of the proposed distributed algorithm. 

\end{abstract}

\section{INTRODUCTION}
\label{SEC1}
Autonomous vehicles are part of our daily lives~\cite{IHSMarkit}.  
These technologies have the potential to reduce fatalities, improve transportation efficiency, and enhance the overall quality of life~\cite{Fagnant2015}. For example, automated vehicles can alleviate the workload of truck drivers on long drives on busy highways and improve efficiency, by forming platoons with neighboring vehicles. Compared to human-driven vehicles, autonomous vehicles are equipped with sensors that allow them to explicitly communicate with each other and share their intentions. To fully exploit the benefits that these technologies will bring, algorithms to control and efficiently coordinate these autonomous systems are crucial, especially in tight environments (e.g., busy highway lanes or parking lots), where the vehicles have reduced maneuvering space. Coordination can be implemented in a centralized or distributed fashion. Having a central coordinator in the context of autonomous driving, however, is an unrealistic assumption if one considers the scale of the road network and the number of vehicles on the road. Autonomous vehicles should be able to coordinate independently in a distributed fashion to avoid the use of a central coordinator to supervise their interactions. To reach their individual goals, each vehicle should closely interact with its neighbors to avoid collisions. Hence, one of the challenges to ensuring the safe navigation of connected autonomous vehicles is that of efficiently generating a safe coordination strategy in a distributed manner. This is a multi-vehicle local trajectory optimization problem. 

In this work, we focus on efficient, safe, and coordinated multi-vehicle distributed trajectory optimization (the interested reader can refer to \cite{Yan2013} for an overview of the state of the art). In particular, we consider the application of our proposed approach for multi-lane tight platooning. This application requires tight maneuvering as discussed in \cite{Firoozi2021}.  Platooning in a classical setting refers to a group of vehicles forming a road train in a single lane \cite{Hedrick1991}, \cite{Shladover1991}. A drawback of single-lane platoons is that a long train-like formation may impede other vehicles from changing lanes, impacting traffic flow and reducing mobility. Platoon formation in multiple lanes combines the advantages of platooning while being shape-reconfigurable and facilitating lane changes. Introducing another degree of freedom in multi-lane platoons increases structural flexibility and can enhance mobility. While single-lane platooning (one-dimensional 1D) is well-studied in the literature, research on multi-lane platoons (two-dimensional 2D) is limited.
\begin{figure}[t]
\centering
\includegraphics[width=\columnwidth]{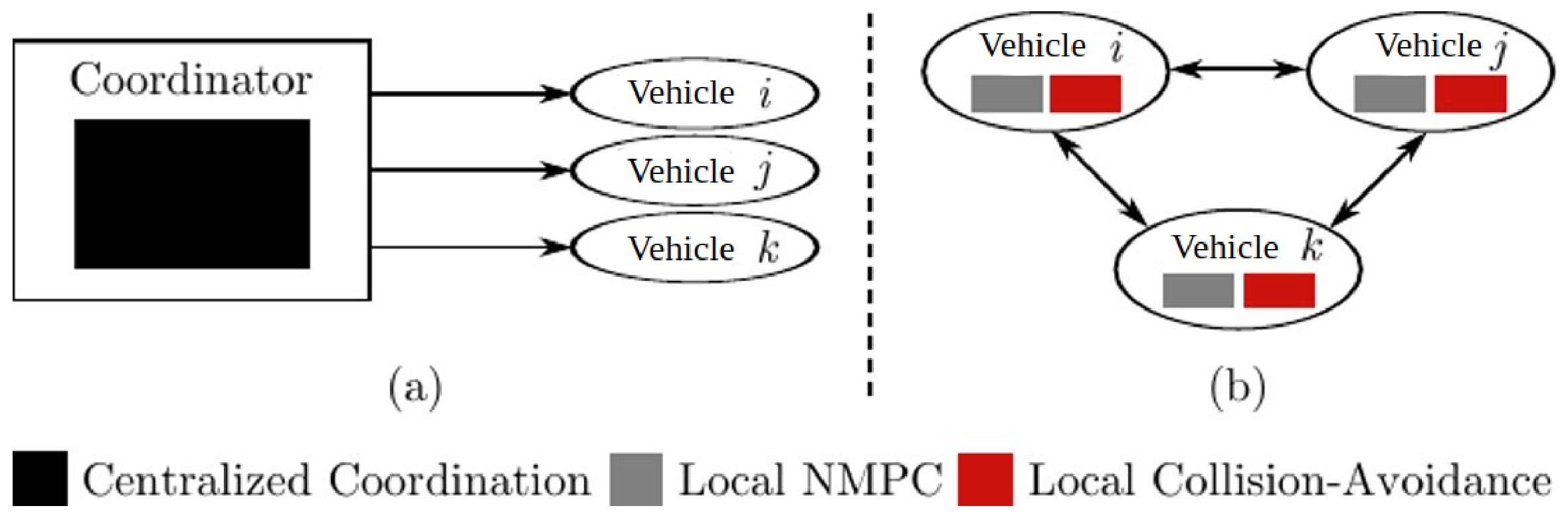}
\caption{Centralized design (a) vs. our distributed design (b) for multi-vehicle coordination.}
\label{fig:cen_decen_comparison}
\end{figure}

In one of our prior works \cite{Firoozi2021}, multi-lane platooning in a centralized way is presented. Although the centralized approach provides tight maneuvering for cooperative vehicles, the centralized framework does not provide computational efficiency for online applications, and the trajectories are all computed offline. In this work, we present a distributed framework that not only provides tight maneuvering but also is computationally efficient for online applications. Compared to the existing literature, to the best of our knowledge, this work is the first one that considers the exact formulation of collision avoidance with the polytopic representation of vehicles in a distributed framework that provides cooperative tight maneuvering of multiple vehicles with computation efficiency suitable for online applications. We consider multi-lane platooning as an application that requires real-time performance as well as tight maneuvering. Tight maneuvering is crucial for multi-lane platoons since the vehicles move in close proximity to each other and need to be able to perform cooperative lane-change at high speed in small spaces in real-time.
In this work, we consider cooperative multi-lane merging of vehicles as a challenging case study to show the effectiveness of our proposed algorithm. However, our proposed algorithm is generalizable to other scenarios in the autonomous driving setting. For example, cooperative autonomous parking in tight environments can benefit from the proposed approach.

\subsection{Contribution}
We use a nonlinear model predictive control (NMPC) to plan collision-free trajectories to coordinate the vehicles. We use a polytopic representation of the individual vehicle and formulate the collision avoidance problem as the problem of finding the minimum distance between two polytopes. This polytopic representation is particularly suitable to precisely represent vehicles, such as cars and vessels, when the maneuver space is limited, for example on busy lanes, parking lots, and canals. To incorporate this collision avoidance strategy in the NMPC formulation, our method relies on duality theory~\cite{Zhang2018}. We reformulate the minimum distance collision-avoidance constraints between each pair of vehicles as a feasibility test (with associated collision-avoidance variables) that can be included within the constraints of the NMPC problem. Solving the NMPC problem in a centralized way (Fig.~\ref{fig:cen_decen_comparison}.a) is computationally intensive, so we introduce an algorithm to solve the NMPC problem in a distributed way (Fig.~\ref{fig:cen_decen_comparison}.b), which is computationally efficient compared to the centralized formulation and is suitable for real-time applications.   


In order to split the centralized problem into distributed sub-problems, the centralized formulation must be separable. Although the dynamic models of the vehicles are decoupled, the collision avoidance constraints are coupled among them.
To break the coupling, we rely on an alternative optimization approach to decompose the centralized problem as local minimization problems performed by alternating between two different optimizations (Fig.~\ref{fig:cen_decen_comparison}.b): \emph{(i)} a collision avoidance optimization (red boxes in Fig.~\ref{fig:cen_decen_comparison}.b) that computes the predicted collision-avoidance variables, given the latest predicted intention of each pair of vehicles, and \emph{(ii)} local NMPC optimizations (gray boxes in Fig.~\ref{fig:cen_decen_comparison}.b) that update the vehicle states, given the latest predicted collision-avoidance variables. The advantage of this decomposition is that each collision avoidance optimization solves efficiently convex problems of fixed dimension and the local NMPC problems have always a fixed number of decision variables (the local vehicle states), compared to the centralized problem.  
Finally, we validate our method for the autonomous lane-merging scenario, where a team of connected autonomous vehicles coordinate on a highway to form a platoon, comparing its performance with a centralized implementation. In lane-merging scenarios both road geometry and formation geometry restrict the motion of the vehicles. Hence, the vehicles must coordinate in a \emph{tight} environment. To allow navigation at tight spaces, our approach models the road structure and the vehicle dimensions, as exact sizes with no approximation or enlargement. Our contributions are summarized as follows:
\begin{itemize}
    \item A distributed alternating optimization scheme using NMPC and dual decomposition for multi-vehicle coordination that provides tight maneuvering.
    \item A collision avoidance reformulation using strong duality theory that allows our method to exploit the dual structure to split the large centralized coordination problem into smaller sub-problems, which are computationally efficient for online applications. 
    \item A geometric interpretation of dual variables. 
\end{itemize}

\subsection{Related Work} 
\subsubsection{Multi-Vehicle Coordination} Classical methods for multi-vehicle coordination either use reactive strategies (such as potential fields~\cite{Schneider2003,Tanner2005,Gayle2009}, dynamic window~\cite{Fox1997}, and velocity obstacles~\cite{Fiorini1998,VanDenBerg2011}), assume a priority order~\cite{Cap2015}, or rely on a scheduling~\cite{Bruni2013} for the vehicles. These methods, however, do not explicitly consider the interaction among the vehicles. Learning-based methods~\cite{Arai1999,Bu2008,Guo2010,La2015,Chen2017} and constrained-optimization approaches can be used to take these interactions into account. Our work fits in this last category and relies on tools from control and optimization to model the interactions among the vehicles to avoid collisions. Distributed constrained-optimization designs have been proposed for example in \cite{Keviczky2007,VanParys2017,Alonso2018,Chen2018,Rey2018,ferranti2022distributed}. The authors in~\cite{Keviczky2007} present a decentralized model predictive control (MPC) formulation for multi-vehicle coordination that relies on invariant-set theory and mix-integer linear programming (MILP). The authors in~\cite{VanParys2017} propose a distributed MPC design for formation control using the alternating direction method of multipliers (ADMM) and separating hyperplanes for collision avoidance. The authors in~\cite{Alonso2018} use a potential cost function and collision-avoidance constraints to formulate a distributed MPC problem, in which the collision avoidance constraints can be either linearized or formulated using integer variables. In addition, the authors rely on motion primitives to account for vehicle kinematic and dynamic constraints. The authors in~\cite{Chen2018,Rey2018} present distributed MPC approaches that rely on ADMM to decompose the (linearized) coordination problem. The authors in~\cite{ferranti2022distributed} propose a distributed nonlinear MPC formulation with nonconvex collision avoidance constraints.

Compared to ~\cite{Keviczky2007}, our proposed approach does not require the solution of a MILP problem that can be computationally expensive to solve. In addition, compared to ~\cite{Alonso2018, Chen2018, Rey2018} our method does not require any linearization (which could reduce the solution space of the problem) of the collision-avoidance constraints. Also compared to ~\cite{Alonso2018}, our approach does not require the use of motion primitives (the vehicle dynamics are directly included in the NMPC formulation). Compared to ~\cite{VanParys2017}, our strategy allows to specify a desired distance between the vehicles, instead of using separating hyperplanes. Inspired by~\cite{Zhang2018, ZhangParking2018}, our method uses dual optimization to formulate the collision avoidance constraints. Compared to~\cite{Zhang2018, ZhangParking2018}, however, our method exploits the structure of the coordination problem to solve it in a distributed fashion. Compared to~\cite{ferranti2022distributed}, we rely on the structure of the dual problem to deal with collision avoidance constraints and on a polytopic representation of the vehicles. 
In \cite{9760054}, a vehicle optimal overtaking maneuver planning is studied. A two-layer trajectory optimization framework is introduced and a near-optimal reference generator is presented. 
While this work considers trajectory optimization for overtaking maneuvers, our work presents a distributed optimization for the simultaneous optimization of multiple cooperative vehicles.

\subsubsection{Outline} Sec.~\ref{SEC2} provides the required preliminary definitions. Sec.~\ref{SEC3} describes the centralized design. Sec.~\ref{SEC4} details our distributed algorithm. Sec.~\ref{SEC5} introduces our applications and the simulation results. Sec.~\ref{SEC7} concludes this paper.

\section{PRELIMINARIES}
\label{SEC2}
We provide the needed definitions and notations below.
\subsubsection{Vehicles and Neighbor vehicles} The set of $M$ cooperative vehicles is defined as $\mathcal V := \{1,2,...,M\}$. We identify each vehicle through its index $i \in\mathcal V$. Throughout this paper, the superscript $i$ denotes the $i$th vehicle. The neighbor set of vehicle $i$ is denoted as $\mathcal{N}_i$ and represents all the vehicles that are in the communication range of vehicle $i$.
\subsubsection{Polytopic Description of Vehicle} The region occupied by the vehicle can be described as a convex set defined by a polytope $\mathcal{P}$. Polytopes are described as the intersection of a set of half-spaces and are defined as a set of linear inequalities. The initial polytopic set of the vehicle is represented as $\mathcal{P}_{o}$. As the vehicle travels, $\mathcal{P}_{o}$ undergoes affine transformations including rotation and translation. Hence $\mathcal{P} = \mathbf{R}\mathcal{P}_{o} + \mathbf{t}_r$, where $\mathbf{R}:\mathbb{R}^{n_{z}} \rightarrow \mathbb{R}^{n\times n}$ is an orthogonal rotation matrix, $\mathbf{t}_r:\mathbb{R}^{n_{z}} \rightarrow \mathbb{R}^{n}$ is the translation vector, $n_z$ is the dimension of the vehicle state $\mathbf{z}$, and $n$ is dimension of the space which is 2 for two-dimensional (2D) planning. Since the rotation matrix and translation vector are functions of vehicle states, the vehicle's polytopic representation $\mathcal{P}$ is a time-varying polytope $\mathcal{P}(\mathbf{z}(t))$.
\subsubsection{Dynamic Obstacles}
$\mathcal{N}_i$ is the set of dynamic obstacles for vehicle $i$. To avoid collision between vehicles $i$ and $j$, the intersection of their polytopic sets must be empty, $\mathcal{P}^i\cap \mathcal{P}^j = \emptyset$.
\subsubsection{MPC Scheme} MPC is useful for online local motion planning in dynamic environments because it is able to re-plan according to the new available information. MPC relies on the receding-horizon principle. At each time step it solves a constrained optimization problem and obtains a sequence of optimal control inputs that minimize a desired cost function $J$, while considering dynamic, state, and input constraints, over a fixed time horizon. Then, the controller applies, in closed loop, the first control-input solution. At the next time step, the procedure is repeated. Throughout this paper, $(\cdot|t)$ indicates the values along the entire planning horizon $N$, predicted based on the measurements at time $t$. For example $\mathbf{z}(\cdot|t)$ represents the entire state trajectory along the horizon $[\mathbf{z}(1|t),\mathbf{z}(2|t),...,\mathbf{z}(N|t)]$ predicted at time $t$. The bar notation ($\mathbf{\bar{\cdot}}$) represents constant known values.
\subsubsection{MPC Cost Function $J$} The MPC cost is $J := \sum_{i\in \mathcal V} J^i$, where  $J^i$ is the local objectives of each vehicle. Each $J^i$ can be designed according to the local vehicle planning and control objectives. For example, the local costs can be specified to reach a goal set or to reduce the deviation from a global reference path (which is not collision-free) computed using high-level planning methods.

\section{CENTRALIZED COORDINATION}
\label{SEC3}
The multi-vehicle coordination can be considered as a motion-planning problem and formulated as a centralized MPC optimization problem that computes collision-free trajectories for all the vehicles, simultaneously. The optimization problem is formulated in the NMPC framework as follows 
\begin{subequations}\label{eq:MPC_formulation}
\begin{align}
& \min_{\substack{\mathbf{u}^{i}(\cdot|t)}}
\label{eq:total_cost}
& &\sum_{i=1}^{M}J^i(\mathbf{z}^i,\mathbf{u}^i) \\
\label{eq:dynamic_constraint}
&\textrm{subject to} & & \mathbf{z}^i(k+1|t) = f(\mathbf{z}^i(k|t),\mathbf{u}^i(k|t)),\\
\label{eq:initial_cond}
& & & \mathbf{z}^i(0|t) = \mathbf{z}^i(t),\\
\label{eq:state_input_bound}
& & & \mathbf{z}^i(k|t) \in \mathcal{Z}, \ \mathbf{u}^i(k|t) \in \mathcal{U},\\
\label{collision_avoidance_constraint}
& & & \mathcal{P}(\mathbf{z}^i(k|t))\cap \mathcal{P}(\mathbf{z}^j(k|t)) = \emptyset, \quad  i \neq j\\ 
& & & \forall i \in \mathcal{V},\ j \in \mathcal{N}_i,\notag \nonumber{\textrm{and~}  k \in \{1,2,..,N\}}.
\end{align} \end{subequations}
In the formulation above, $\mathbf{u}^i(\cdot|t) = [u^i(k|t),...,u^i(k+N-1|t)]$ denotes the sequence of control inputs over the MPC planning horizon $N$ for $i$th vehicle. $\mathbf{z}^i(k|t)$ and $\mathbf{u}^i(k|t)$ variables of $i$th vehicle at step $k$ are predicted at time $t$. Also, $\mathbf{u}^i = \{\mathbf{u}^1, ..., \mathbf{u}^M\}$ for $M$ number of vehicles. The function $f(\cdot)$ in \eqref{eq:dynamic_constraint} represents the nonlinear (dynamic or kinematic) model of the vehicle, which is discretized using Euler discretization. $\mathcal{Z}$, $\mathcal{U}$ are the state and input feasible sets, respectively. These sets represent state and actuator limitations. Constraints~\eqref{collision_avoidance_constraint} represent the collision-avoidance constraints between the $i$th vehicle and all the neighboring vehicles within the communication radius. This representation is time-varying and is a function of the vehicle state at each time step. The remainder of this section details the derivation of constraint~\eqref{collision_avoidance_constraint}. 
In this paper, our focus is to reformulate the centralized problem \eqref{eq:MPC_formulation} into a distributed one. Note that safety in this work is referred to hard constraint satisfaction. However, safety guarantees for all times requires guarantees of persistent feasibility of the MPC. Any techniques to guarantee persistent feasibility can be incorporated into our proposed approach since adding terminal cost and terminal set elements (e.g., \cite{kohler2019}) to the formulation will not affect the coupling among the agents. 

\subsection{Collision Avoidance Reformulation}
Consider two polytopic sets $\mathcal{P}_{1}$ and $\mathcal{P}_{2}$. The distance between these sets is given by the following primal problem
\begin{equation}\label{eq:pprimal}
\text{dist}(\mathcal{P}_{1},\mathcal{P}_{2}) =\underset{{\mathbf{x},\mathbf{y}}}{\min}\{\|\mathbf{x}-\mathbf{y}\|_{2}|\mathbf{A}_{1}\mathbf{x} \leq \mathbf{b}_{1}, \mathbf{A}_{2}\mathbf{y} \leq \mathbf{b}_{2}\},\\
\end{equation}
where $\mathcal{P}_{1} = \{\mathbf{x}\in \mathbb R^n|\mathbf{A}_{1}\mathbf{x}\leq \mathbf{b}_{1}\}$ and $\mathcal{P}_{2}=\{\mathbf{y}\in \mathbb R^n|\mathbf{A}_{2}\mathbf{y}\leq \mathbf{b}_{2}\}$. The two sets do not intersect if $\text{dist}(\mathcal{P}_{1},\mathcal{P}_{2}) >0$. For motion-planning applications, the vehicles must keep a minimum safe distance $d_{\min}$ from each other and from the obstacles. Hence, the distance between their polytopic sets should be larger than a predefined minimum distance, 
\begin{equation}\label{eq:dsafe}
    \text{dist}(\mathcal{P}_{1},\mathcal{P}_{2}) \geq d_{\min}.
\end{equation}

Problem \eqref{eq:pprimal} is an optimization problem which is a constraint for optimization problem \eqref{eq:MPC_formulation}. Therefore, using the primal problem \eqref{eq:pprimal} for collision avoidance constraint \eqref{collision_avoidance_constraint} yields a bi-level optimization problem. To reduce compute complexity, we rely on strong-duality theory. Building on~\cite{Zhang2018}, the dual problem can be solved instead of the primal problem \eqref{eq:pprimal}. The dual problem is expressed as follows:
\begin{equation}\label{eq:dual}
\begin{aligned}
\text{dist}(\mathcal{P}_{1},\mathcal{P}_{2}) :=&\underset{{\boldsymbol{\lambda}_{12},\boldsymbol{\lambda}_{21},\,\mathbf{s}}}{\max}
-\mathbf{b}_1^{\top} \boldsymbol{\lambda}_{12} - \mathbf{b}_{2}^{\top} \boldsymbol{\lambda}_{21} \\
&\text{s.t.~} \mathbf{A}_{1}^{\top} \boldsymbol{\lambda}_{12} + \mathbf{s}= 0,\ \mathbf{A}_{2}^{\top} \boldsymbol{\lambda}_{21} - \mathbf{s} = 0,\\ 
&\quad\ ||\mathbf{s}||_2 \leq 1, -\boldsymbol{\lambda}_{12} \leq 0,\ -\boldsymbol{\lambda}_{21} \leq 0,
\end{aligned}
\end{equation} 
where $\boldsymbol{\lambda}_{12}$, $\boldsymbol{\lambda}_{21}$ and $\mathbf{s}$ are dual variables.

The optimal value of the dual problem is the distance between $\mathcal{P}_{1}$ and $\mathcal{P}_{2}$ and based on \eqref{eq:dsafe}, the distance between the two polytopes is constrained to be larger than a desired minimum distance. Consequently, we can use this insight to reformulate the dual problem as the following feasibility problem: 
$\{\exists \boldsymbol{\lambda}_{12} \succeq 0, \boldsymbol{\lambda}_{21} \succeq 0, \mathbf{s}:
-\mathbf{b}_{1}^{\top} \boldsymbol{\lambda}_{12} -\mathbf{b}_2^{\top} \boldsymbol{\lambda}_{21} \geq d_\text{min},
\mathbf{A}_1^{\top} \boldsymbol{\lambda}_{12} + \mathbf{s} = 0,
\mathbf{A}_{2}^{\top} \boldsymbol{\lambda}_{21} - \mathbf{s} = 0,
\|\mathbf{s}\|_2 \leq 1\}.$ This reformulation can be substituted to the collision-avoidance constraint \eqref{collision_avoidance_constraint} in Problem \eqref{eq:MPC_formulation}. 
Therefore, problem \eqref{eq:MPC_formulation} can be rewritten as 
\begin{subequations}\label{eq:MPC_centralized}
\begin{align}
& \min_{\substack{\mathbf{u}^i(\cdot|t),\ \boldsymbol{\lambda}_{ij}(\cdot|t),\\ \boldsymbol{\lambda}_{ji}(\cdot|t),\ \mathbf{s}_{ij}(\cdot|t)}}
& & \nonumber{\sum_{i=1}^{M}J^i(\mathbf{z}^i,\mathbf{u}^i)} \\
&\textrm{subject to} & &
\nonumber{\eqref{eq:dynamic_constraint},\eqref{eq:initial_cond},} \nonumber{\eqref{eq:state_input_bound},}\\
& & & \nonumber{ \big(-\mathbf{b}^i(\mathbf{z}^i(k|t))^{\top}} \boldsymbol{\lambda}_{ij}(k|t) \\
\label{eq:coupling_constraint}
& & & - \mathbf{b}^j(\mathbf{z}^{j}(k|t))^{\top} \boldsymbol{\lambda}_{ji}(k|t)\big) \geq d_\text{min},\\
\label{eq:coupling_sep_hyp_pos}
& & & \mathbf{A}^i(\mathbf{z}^i(k|t))^{\top} \boldsymbol{\lambda}_{ij}(k|t)\! +\!\mathbf{s}_{ij}(k|t)\! =\! 0,\\
\label{eq:coupling_sep_hyp_neg}
& & & \mathbf{A}^j(\mathbf{z}^j(k|t))^{\top} \boldsymbol{\lambda}_{ji}(k|t)\! -\! \mathbf{s}_{ij}(k|t)\!= \!0,\\
& & & \nonumber{\boldsymbol{\lambda}_{ij}(k|t),\boldsymbol{\lambda}_{ji}(k|t) \geq 0,\|{\mathbf{s}_{ij}(k|t)}\|_2 \leq 1,} \\
& & & {\forall i} \nonumber{\in \mathcal{V},\ j \in \mathcal{N}_i,~} \nonumber{\text{and~} k \in \{1,2,..,N\},}
\end{align} 
\end{subequations}
where $\mathbf{A}^i$ and $\mathbf{b}^i$ are functions of $\mathbf{z}^i(k|t)$ and represent the polytopic set of $i$th vehicle at step $k$ predicted at time $t$. Similarly, $\mathbf{A}^j$ and $\mathbf{b}^j$ denote the polytopic set of $j$th vehicle which belongs to neighbor set $\mathcal{N}_i$. The dual variables $\boldsymbol{\lambda}_{ij}$, $\boldsymbol{\lambda}_{ji}$ and $\mathbf{s}_{ij}$ are coupled through the collision avoidance constraint between vehicle $i$ and vehicle $j$.
Note that the dual variable $\mathbf{s}_{ij}$ is equivalent to the variable $\mathbf{s}$ in \eqref{eq:dual}. The new variable $\mathbf{s}_{ij}$ is introduced in \eqref{eq:MPC_centralized} to distinguish for example, $\mathbf{s}_{12}$ from $\mathbf{s}_{13}$, but $\mathbf{s}_{12}$ is identical to $\mathbf{s}_{21}$ according to \eqref{eq:dual}. Therefore, the variables $\mathbf{s}_{ij}$, $\mathbf{s}_{ji}$ and $\mathbf{s}$ are all identical vectors ($\mathbf{s}_{ij}$ = $\mathbf{s}_{ji}$ = $\mathbf{s}$ $\in \mathbb{R}^n$) with dimension of $n$ that is the dimension of the space which is 2 for 2D planning. 
\section{DISTRIBUTED COORDINATION}
\label{SEC4}
Problem~\eqref{eq:MPC_centralized} is a centralized coordination formulation that simultaneously optimizes over all the vehicles' states $\mathbf{z}^i$ and the collision avoidance variables $\boldsymbol{\lambda}_{ij}, \boldsymbol{\lambda}_{ji}, \mathbf{s}_{ij}$ (for all $i=1,...,M$, $j\neq i$), that is, the number of variables to optimize is proportional to the number of vehicles. This is computationally expensive when $M$ is large, making the centralized formulation not scalable with the number of vehicles. Our goal is to remove the need for a central coordinator and make the problem scalable with the number of vehicles (allowing the vehicles to coordinate and locally solve smaller sub-problems in parallel). 

By looking at the structure of Problem~\eqref{eq:MPC_centralized}, we notice that the collision avoidance constraints \eqref{eq:coupling_constraint}-\eqref{eq:coupling_sep_hyp_neg} create a coupling among the vehicles. In addition, Constraint \eqref{eq:coupling_constraint} creates a nonlinear coupling between the state variables $\mathbf{z}^i$,$\mathbf{z}^j$ and the collision avoidance variables $\boldsymbol{\lambda}_{ij}$,$\boldsymbol{\lambda}_{ji}$. To break up these couplings and devise the proposed distributed algorithm, we rely on the dual structure we originally used to formulate Problem~\eqref{eq:MPC_centralized} and on the ability of MPC to generate predictions. In particular, our idea is to solve Problem~\eqref{eq:MPC_centralized} by using an alternating optimization scheme, that is, we replace the central coordinator~\eqref{eq:MPC_centralized} by using two independent optimizations that perform alternating optimization of the dual variables (associated with the collision avoidance constraints) and of primal state variables, respectively, as Algorithm~\ref{alg_coordination} details.
The idea of alternating has a long history in the optimization literature. In particular, the closest approaches to the one presented here are \cite{Rey2018} and \cite{VanParys2017} which use ADMM for distributed collision avoidance. The core difference between these approaches and ours is that our alternating optimization scheme is based on reformulation of collision avoidance using duality theory and exploiting the dual structure of the problem for decomposition. Also, our approach is based on polytopic representation of the vehicles which facilitates navigation in tight environments by modeling the exact dimensions of the vehicle without over-approximation. However, the previous studies  \cite{Rey2018} consider the vehicle shape as a simple circle, which is not advantageous where the environment is tight.  

In Algorithm~\ref{alg_coordination}, first, the dual variables over the NMPC horizon are initialized, then the first optimization step (NMPC optimization) optimizes the state variables $\mathbf{z}^i$,$\mathbf{z}^j$ over the horizon while keeping the dual variables $\boldsymbol{\lambda}_{ij}$,$\boldsymbol{\lambda}_{ji}, \mathbf{s}_{ij}$ fixed. The second optimization step (Collision Avoidance, CA optimization) optimizes the dual variable while keeping the state variables fixed. We detail these two optimizations below.
\subsection{NMPC optimization}
At time $t$, each vehicle $i$ independently computes its own state $\mathbf{z}^i$ trajectory, given the dual variables over the horizon $[\boldsymbol{\lambda}_{ij}(1)$, $\dots$, $[\boldsymbol{\lambda}_{ij}(N)]$, $[\boldsymbol{\lambda}_{ji}(1)$, $\dots$, $\boldsymbol{\lambda}_{ji}(N)]$ and $[\mathbf{s}_{ij}(1)$, $\dots$, $\mathbf{s}_{ij}(N)]$. For each vehicle $ i \in \mathcal{V},\ j \in \mathcal{N}_i$, the NMPC optimization is given by:
\begin{subequations}\label{eq:MPC_optimization}
\begin{align}
& \min_{\substack{\mathbf{u}^i(\cdot|t)}}
& & \nonumber{J^i(\mathbf{z}^{i},\mathbf{u}^{i})}\\
& \textrm{subject to} & &
\nonumber{\eqref{eq:dynamic_constraint},\eqref{eq:initial_cond}, \eqref{eq:state_input_bound}},\\
& & & \nonumber{\big(-\mathbf{b}^{i}(\mathbf{z}^i(k|t))^{\top} \boldsymbol{\bar\lambda}_{ij}(k|t)}\\ 
& & & - \mathbf{\bar b}^j(\mathbf{\bar z}^j(k|t))^{\top} \boldsymbol{\bar \lambda}_{ji}(k|t)\big) \geq d_\text{min}, \label{NMPC_dmin}\\
& & & \mathbf{A}^{i}(\mathbf{z}^i(k|t))^{\top} \boldsymbol{\bar \lambda}_{ij}(k|t) + \mathbf{\bar s}_{ij}(k|t) = 0, \label{distributed_sij}\\
& & & \nonumber{\text{for all } k \in \{1,2,..,N\}},
\end{align} \end{subequations}
where the bar notation ($\mathbf{\bar{\cdot}}$) represents constant known values and $\mathbf{A}^i$, $\mathbf{b}^i$ are the polytopic representation of the $i$th vehicle and are functions of $\mathbf{z}^i$. The optimized trajectory is then shared with the collision avoidance optimization (shifted in time according to step \textcircled{\footnotesize{5}} of Algorithm~\ref{alg_coordination} to account for the 1-step delay in the calculation of the collision-avoidance strategies). The last step of the shifted trajectory, denoted as $\mathbf{z}_{\textrm{IT}}$ is interpolated by assuming constant velocity and evolving the dynamics one step forward from the state $\mathbf{z}(N)$. Problem~\eqref{eq:MPC_optimization} can be solved in parallel by each vehicle. In this optimization, the collision-avoidance variables $\boldsymbol{\lambda}_{ij}, \boldsymbol{\lambda}_{ji}, \mathbf{s}_{ij}$ are considered as known values along the planning horizon. 
Note that compared to the centralized formulation, the only decision variable to optimize in the NMPC optimization is the $i$th vehicle state $\mathbf{z}^i$ (i.e., the number of decision variables in the local problem formulations is constant). 
\subsection{Collision Avoidance (CA) optimization}
Each vehicle pair of $ i, j \in \mathcal{N}, \ i \neq j$, computes the collision avoidance variables $\boldsymbol{\lambda}_{ij}, \boldsymbol{\lambda}_{ji}, \mathbf{s}_{ij}$. The CA optimization is given by 
\begin{subequations}\label{eq:collision_optimization}
\begin{align}
& \max_{\substack{\boldsymbol{\lambda}_{ij}(\cdot|t),\\ \boldsymbol{\lambda}_{ji}(\cdot|t),\\ \mathbf{s_{ij}}(\cdot|t)}}
&&\!\!\!\!\!\!\!\!\! -\! \nonumber{\mathbf{\bar b}^{i}\!(\mathbf{\bar{z}}^i(k|t))^{\top}\! \boldsymbol{\lambda}_{ij}(k|t)\!-\! \mathbf{\bar b}^{j}\!(\mathbf{\bar{z}}^j(k|t))^{\top}\! \boldsymbol{\lambda}_{ji}(k|t)}\\
& \textrm{subject to} & &
 \mathbf{\bar A}^{i}(\mathbf{\bar{z}}^{i}(k|t))^{\top} \boldsymbol{\lambda}_{ij}(k|t) + \mathbf{s}_{ij}(k|t) = 0, \label{CA_lambda_ij}\\
& & &\mathbf{\bar A}^{j}(\mathbf{\bar{z}}^{j}(k|t))^{\top} \boldsymbol{\lambda}_{ji}(k|t) - \mathbf{s}_{ij}(k|t) = 0, \label{CA_lambda_ji}\\
& & & \nonumber{\big(-\mathbf{\bar b}^{i}(\mathbf{\bar z}^i(k|t))^{\top} \boldsymbol{\lambda}_{ij}(k|t)}\\
& & & \nonumber{-\mathbf{\bar b}^j(\mathbf{\bar z}^j(k|t))^{\top} \boldsymbol{\lambda}_{ji}(k|t)\big) \geq d_\text{min},}\\
& & &\|{\mathbf{s}_{ij}(k|t)}\|_2 \leq 1, \\
& & & \nonumber{-\boldsymbol{\lambda}_{ji}(k|t)\! \leq\! 0, -\boldsymbol{\lambda}_{ij}(k|t) \leq 0,}\\
& & & \nonumber{\text{for all } k \in \{1,2,..,N\}.}
\end{align} \end{subequations}
Each vehicle solves Problem \eqref{eq:collision_optimization} in parallel. This optimization assumes the state trajectories of the vehicles $\mathbf{z}^i$ to be fixed (obtained by the NMPC optimization and from the neighboring vehicles according to step \textcircled{\footnotesize{7}} of Algorithm~\ref{alg_coordination}). This problem can be solved efficiently.
Note that the decision variables of this optimization are dual variables that are shared between a pair of vehicles. However, each vehicle solves this problem for itself independently. A pair of vehicles $i$ and $j$ solve the same problem and find the same solutions for shared dual variables $\boldsymbol{\lambda}_{ij}, \boldsymbol{\lambda}_{ij}$ and $\mathbf{s_{ij}}$, so the approach is fully distributed, which means each individual vehicle solves its own planning problem independently.    

Algorithm \ref{alg_coordination} is an alternating optimization scheme with the NMPC optimization \eqref{eq:MPC_optimization} and the CA optimization \eqref{eq:collision_optimization}. On one hand, this alternating optimization scheme allows us to improve the computation time of the coordination strategy. On the other hand, due to this optimization scheme, the distributed NMPC \eqref{eq:MPC_optimization} returns less tight trajectories (larger margins in the coordination) compared to the centralized NMPC \eqref{eq:MPC_centralized}. Solving CA optimization and substituting its solution in the NMPC optimization further restricts the constraints \eqref{NMPC_dmin}-\eqref{distributed_sij}, since the dual variables are kept fixed (i.e., the NMPC optimizer has less degree of freedom in the computation of the local trajectories). In contrast, in the centralized NMPC \eqref{eq:MPC_centralized}, the equivalent constraints \eqref{eq:coupling_constraint}-\eqref{eq:coupling_sep_hyp_neg} can be interpreted as the relaxed version of the constraints \eqref{NMPC_dmin}-\eqref{distributed_sij} since the dual variables are decision variables of the centralized problem.
\begin{algorithm}[t]
\caption{\small Distributed Coordination Algorithm}\label{alg_coordination}
    \begin{algorithmic}[1]
    \State This algorithm alternates between two optimizations, \textbf{NMPC Optimization \eqref{eq:MPC_optimization}} and \textbf{Collision Avoidance Optimization \eqref{eq:collision_optimization}}.
    \State Initialize $[\mathbf{s}_{ij}(1),...,\!\mathbf{s}_{ij}(N)],$ $[\boldsymbol{\lambda}_{ij}(1),...,\boldsymbol{\lambda}_{ij}(N)],$ $[\boldsymbol{\lambda}_{ji}(1),...,\boldsymbol{\lambda}_{ji}(N)]$, $\forall i,j \in \mathcal{N}$  and $i\neq j.$
        \For { $t=0,1,..., \infty$} 
        \ForAll {vehicle $i$, $i\in \mathcal V$} \Comment{in parallel}
          \State {Solve Problem~\eqref{eq:MPC_optimization}}
        \State {Compute the shifted state and interpolate the last state \quad \quad $[\mathbf{z}^i(2),...,\mathbf{z}^i(N),\mathbf{z}^i_{\textrm{IT}}]$.}
        \State  {Compute the associated polytopic sets: $[\mathbf{A}^i(2),...,$ \hphantom{zzzzzzz}$\mathbf{A}^i(N),$ $\mathbf{A}^i_{\textrm{IT}}]$, $[\mathbf{b}^i(2),$..., $\mathbf{b}^i(N),$ $\mathbf{b}^i_{\textrm{IT}}].$}
        \State {Communicate to vehicle $j$ ($\forall j \in \mathcal N_i$) the polytopic sets.}
        \State {Solve Problem \eqref{eq:collision_optimization} for each $j \in \mathcal{N}_i$}
        \State {Apply $\mathbf{u}_{\text{MPC}}^{i}$ to move forward.}
        \EndFor
        \EndFor
    \end{algorithmic}
\end{algorithm}

\subsection{Geometric Interpretation of Primal and Dual Variables}
The dual variables have an interesting geometric interpretation. All these geometric meanings are obtained from the Karush–Kuhn–Tucker (KKT) conditions for problem \eqref{eq:pprimal} (more details can be found in ~\cite{DAX2006}). Since the primal problem is a convex function, the KKT condition is necessary and sufficient for the points to be primal and dual optimal. 
The derivation of the KKT conditions upon which the geometric interpretations are obtained are provided in the Appendix. 
As seen in \figref{fig:geometric_rep}, the top plots show the geometric representation of the primal formulation \eqref{eq:pprimal} in which the optimal solutions are $\mathbf{x}^*$ and $\mathbf{y}^*$ and the distance is defined as the \emph{classical} Euclidean distance $\|\mathbf{x}^*-\mathbf{y}^*\|_2$ between the two sets. The bottom plots show the equivalent dual formulation \eqref{eq:dual} in which the optimal solutions are $\mathbf{s}^*$, $\boldsymbol{\lambda}_{12}^*$ and $\boldsymbol{\lambda}_{21}^*$ and the same distance between the two polytopic sets is defined as $-{\mathbf{b}_{1}}^{\top} \boldsymbol{\lambda}_{12}^* - {\mathbf{b}_{2}}^{\top} \boldsymbol{\lambda}_{21}^*$. As \figref{fig:geometric_rep}(e) depicts, separating the hyperplane between the two polytopic sets is always perpendicular to the minimum distance. Therefore $\mathbf{s}^*$, which is the normal vector of a separating hyperplane, is always parallel to the minimum distance. The normal vector $\mathbf{s}^*$ plays the role of the consensus variable between the vehicles. As discussed earlier, the vector $\mathbf{s}^*$ is shared between each pair of vehicles according to \eqref{eq:dual}, so $\mathbf{s}_{ij} = \mathbf{s}_{ji}$. Furthermore, in \figref{fig:geometric_rep}(f), the green lines show two supporting hyperplanes which are parallel to the separating hyperplane. The hyperplane $\mathbf{s}^*{^{\top}}\mathbf{x} = -\mathbf{b}_1^{\top} \boldsymbol{\lambda}^*_{12}$ supports the set $\mathbf{x}$ or $(\mathcal{P}_1)$ at the point $\mathbf{x}^*$. Similarly the hyperplane $\mathbf{s}^*{^{\top}}\mathbf{y} = \mathbf{b}_2^{\top} \boldsymbol{\lambda}^*_{21}$ supports the set $\mathbf{y}$ or $(\mathcal{P}_2)$ at the point $\mathbf{y}^*$. The primal \eqref{eq:pprimal} and dual \eqref{eq:dual} problems are convex and Slater's condition is satisfied, so the strong duality holds \cite{BoydBook2004}. Therefore, finding the shortest distance between two polytopic sets (primal problem \eqref{eq:pprimal}) is equivalent to finding the maximal separation, which is the maximum distance between a pair of parallel hyperplanes that supports the two sets (dual problem \eqref{eq:dual}) as shown in \figref{fig:geometric_rep}(f).
\begin{figure}
\centering
\includegraphics[width=\columnwidth]{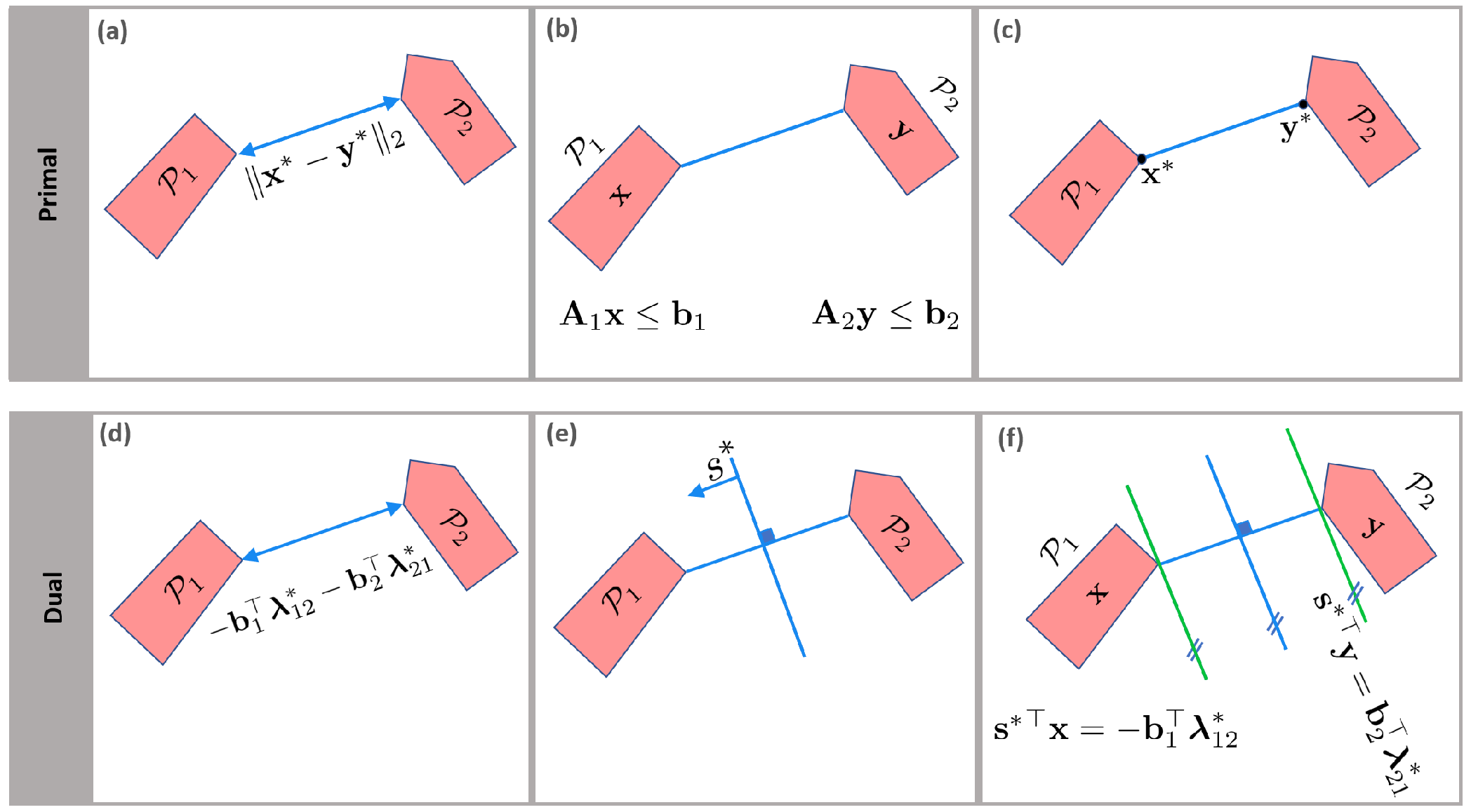}
\caption{\textbf{Top: Primal} (a) minimum distance (b) polytope representation (c) optimal solutions. \textbf{Bottom: Dual}: (d) minimum distance (e) separating hyperplane (f) supporting hyperplanes.} 
\label{fig:geometric_rep}
\end{figure}

\section{NUMERICAL RESULTS FOR AUTONOMOUS DRIVING APPLICATION}
\label{SEC5}
In this section, we conduct a comparative analysis of the performance between centralized and distributed approaches, evaluating their computation efficiency, scalability, and capability for tight maneuvering. Our focus lies in the context of multi-lane tight platooning, wherein vehicles from different lanes merge to achieve a desired configuration. The viability of a cooperative lane merging scenario is assumed to be assessed by a high-level Traffic Operation System (TOS). The TOS monitors and predicts surrounding traffic, offering cooperative vehicles a desired platoon configuration or formation pattern. It's important to note that the design of the TOS decision-making system is beyond the scope of our work. Our emphasis is on motion planning for multiple vehicles, assuming that the surrounding traffic permits the formation or reconfiguration of the platoon. By concentrating on these aspects, we aim to shed light on the comparative strengths and limitations of centralized and distributed approaches in the specific context of multi-lane tight platooning.

In Section~\ref{assumptions} the assumptions are provided. Then Section~\ref{subsec:sim_setup} describes the simulation setup. Section~\ref {subsec:platoon_formation} presents a platoon formation scenario. This Section includes NMPC quantities, high-level reference generator, vehicle polytopic shapes, numerical results, and performance evaluations and discussion.

\subsection{Assumptions}\label{assumptions}
In this work, the following assumptions have been made:
\begin{itemize}
\item[(A1)] The vehicles are fully autonomous and connected through vehicle-to-vehicle (V2V) and vehicle-to-cloud (V2C) communications.
\item[(A2)] The desired platoon configuration $\mathcal{C}_f$ is available from a higher-level traffic operation system (TOS). The vehicles communicate with TOS via V2C communication to determine if based on the current surrounding traffic a platoon formation is possible to not. 
\item[(A3)] Uncertainty due to communication delay or model mismatch is not considered.
\end{itemize}

\subsection{Simulation Setup}
\label{subsec:sim_setup}
We tested our design on a quad-core CPU Intel Core i7-7700HQ @ 2.80 GHz in MATLAB using the MATLAB Parallel Computing Toolbox to simulate the individual vehicles and communication exchanges among the vehicles. We modeled the optimization problems in YALMIP~\cite{Lofberg2004}. We solved Problems~\eqref{eq:MPC_centralized},~\eqref{eq:MPC_optimization} using IPOPT~\cite{Wachter2006}, a state-of-the-art interior-point solver for non-convex optimization, and we solved Problem~\eqref{eq:collision_optimization} using Gurobi~\cite{gurobi}, an efficient QP problem. 
For all the scenarios the simulation results are presented as top-view snapshots, as well as a series of state and action plots. The vehicle colors of the snapshots and plots are matched. 
\subsection{Platoon Formation}
\label{subsec:platoon_formation}
In this scenario, autonomous and connected vehicles merge into a platoon (train-like formation) and maintain a close inter-vehicular distance within the group. In platooning on public roads, both road geometry (lane width) and platoon geometry (longitudinal and lateral inter-vehicle spacing) restrict the motion of the vehicles within the platoon. Hence, the vehicles must coordinate in a \emph{tight} environment. To allow navigation in tight spaces, it is essential to model the road structure and the vehicle dimensions, including length $h$ and width $w$, as exact sizes with no approximation or enlargement. 
\subsubsection{Relevant NMPC Quantities}
\label{subsec:vehicle_model}
We model each vehicle $i$ within the platoon by using a nonlinear kinematic bicycle model (a common modeling approach in path planning) described by the following equations~\cite{Kong2015}.
(we omit the superscript $i$ when it is clear from the context): 
\begin{equation}\label{eq:kinematic_bicycle_model}
\begin{aligned}
\dot x & = v \cos(\psi+\beta),\
\dot y  = v \sin(\psi+\beta), \\
\dot \psi &= \frac{v\cos\beta}{l_f+l_r}(\tan\delta),\
\dot v = a,
\end{aligned}
\end{equation}
where the $i$th vehicle state vector is $\mathbf{z} = [x,y,\psi,v]^\top$ ($x$, $y$, $\psi$, and $v$ are the longitudinal position, the lateral position, the heading angle, and the velocity, respectively)
, the control input vector is $\mathbf{u} = [a, \delta]^\top$ ($a$ and $\delta$ are the acceleration and the steering angle, respectively), $\beta := \arctan\big(\tan\delta(\frac{l_r}{l_f+l_r})\big)$ is the side slip angle, and $l_{f}$, $l_{r}$ are the distance from the center of gravity to the front and rear axles, respectively.  Using Euler discretization, the model \eqref{eq:kinematic_bicycle_model} is discretized with sampling time $\Delta t$ as
\begin{equation}\label{eq:kinematic_bicycle_model_discrete}
\begin{aligned}
x(t+1) & = x(t) + \Delta t \ v(t) \cos(\psi(t)+\beta(t)),  \\
y(t+1) & = y(t)+\Delta t \ v(t)  \sin(\psi(t)+\beta(t)), \\
\psi(t+1) &= \psi(t)+ \Delta t \ \frac{v(t)\cos\beta(t)}{l_f+l_r}(\tan\delta(t)),\\
v(t+1) &= v(t) + \Delta t \ a(t).
\end{aligned}
\end{equation}
The local costs are defined as 
\begin{equation}\label{eq:cost}
\begin{aligned}
J(\mathbf{z},\mathbf{u})&= \sum_{k=t}^{t+N}\|(\mathbf{z}(k|t)-\mathbf{z}_\textrm{Ref}(k|t))\|^2_{Q_\mathbf{z}}\\&
+ \sum_{k=t}^{t+N-1}(\|(\mathbf{u}(k|t))||^2_{Q_\mathbf{u}} + \|(\Delta \mathbf{u}(k|t))\|^2_{Q_{\boldsymbol{\Delta u}}}),
\end{aligned}
\end{equation}
where $\Delta \mathbf{u}$ penalizes changes in the input rate.
$Q_\mathbf{z}\succeq 0$, $Q{_\mathbf{u}}, Q_{\boldsymbol{\Delta u}}\succ 0$ are weighting matrices, $\mathbf{z}_\textrm{Ref}$ is the reference trajectory generated by a high-level planner. Note that this reference trajectory is not collision-free. 
\subsubsection{Reference Generator Model} To generate the reference trajectories for each vehicle, a simple integrator function is defined. The function $g: \mathbb R^3 \to \mathbb R^T,$ is defined as
\begin{equation}\label{eq:h}
g: (x(0),v_{\text{sp}},T) \to 
x_{\text{Ref}} = [x(0), x(1),...,x(T)],
\end{equation} 
which determines $x_{\text{Ref}}$ for all the vehicles within the platoon. The trajectory is obtained by $x(t+1) = x(t) + v_{\text{sp}}\Delta t,\quad \forall{t} \in \{0,1,...,T\}$, where $x(t)$ is the vehicle longitudinal position at time $t$, $v_{\text{sp}}$ is the maximum speed limit of the road, $T$ is the final time of simulation and $\Delta t$ is the simulation sampling time.

\subsubsection{High-Level Reference Generation}
\begin{figure}[ht]
\centering
\includegraphics[scale = 0.8]{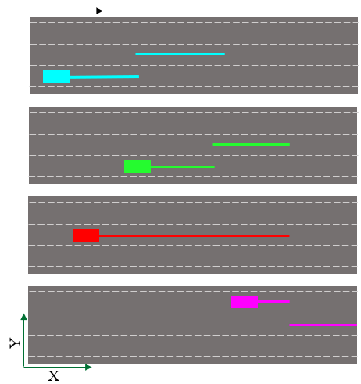}
\caption{The generated reference trajectories are shown for the example scenario of \figref{fig:snapshot_4vehivle}. The parameter $\rho = 0.5$ for the cyan and green vehicles.}
\label{fig:ref_generation}
\end{figure}

The reference state for $i$-th vehicle is denoted as $\mathbf{z}^i_{\text{Ref}} = [x^{i}_{\text{Ref}},y^{i}_{\text{Ref}},\psi^{i}_{\text{Ref}},v^{i}_{\text{Ref}}]$. The reference state trajectory, denoted as $\boldsymbol{\tau}^i_{\mathbf{z}_{\text{Ref}}}$, is defined for the interval $[0,1,2,\hdots,T]$, from the initial time $0$ until the final maneuver time $T$ and $\boldsymbol{\tau}^i_{\mathbf{z}_{\text{Ref}}}$ = $\{\mathbf{z}^i_{\text{Ref}}(0), \mathbf{z}^i_{\text{Ref}}(1), \mathbf{z}^i_{\text{Ref}}(2),\hdots, \mathbf{z}^i_{\text{Ref}}(T)\}$. It is computed based on the initial position of the vehicles $(x(0),y(0))^i \quad \forall{i} \in \mathcal{V}$ and final desired configurations of the platoon $\mathcal{C}_{f}$. The longitudinal position reference trajectory $\boldsymbol{\tau}^i_{x_{\text{Ref}}}$ = $\{x^{i}_{\text{Ref}}(0),\hdots, x^{i}_{\text{Ref}}(T)\}$ is generated using the integrator model \eqref{eq:h},
\begin{equation}\label{x_ref}
    \boldsymbol{\tau}^i_{x_{\text{Ref}}} = g(x^i(0),v_{\text{sp}},T).
\end{equation}
The lateral position reference trajectory $\boldsymbol{\tau}^i_{y_{\text{Ref}}}$ is the $y$ coordinate of the road centerline for each vehicle. For the first portion of simulation $(0,\hdots,\rho T)$, $y^{i}_{\text{Ref}}$ is obtained from initial position $(x(0),y(0))^i \, \forall{i}\, \in\, \mathcal{V}$ and the rest $((\rho T+1),\hdots,T)$ is determined by final desired configuration $(x^i(T),y^i(T)) \, \forall{i}\, \in\, \mathcal{V},$
\begin{equation}\label{y_ref}
\{y^{i}_{\text{Ref}}(0),\hdots, y^{i}_{\text{Ref}}(\rho T)\}=y^i(0),
\end{equation}
\begin{equation}\label{y_ref2}
\{y^{i}_{\text{Ref}}(\rho T+1),\hdots, y^{i}_{\text{Ref}}(T)\}=y^i(T),
\end{equation}
the parameter $\rho \in (0,1)$ 
is a tuning parameter, which can be determined by a design engineer.
It is the coefficient that affects the start of the lane change. $\psi^i_{\text{Ref}}$ is zero
\begin{equation}\label{psi_ref}
   \boldsymbol{\tau}^i_{\psi_{\text{Ref}}} = \{\psi^{i}_{\text{Ref}}(0),\hdots, \psi^{i}_{\text{Ref}}(T)\} = 0,
\end{equation}
assuming the road remains straight along the maneuver and $v^{i}_{\text{Ref}}$ is set as the maximum speed limit of the road or average traffic flow $v_{\text{sp}}$.   
\begin{equation}\label{v_ref}
    \boldsymbol{\tau}^i_{v_{\text{Ref}}} = \{v^{i}_{\text{Ref}}(0),\hdots, v^{i}_{\text{Ref}}(T)\} = v_{\text{sp}}.
\end{equation}
The reference trajectory $\boldsymbol{\tau}^i_{\mathbf{z}_{\text{Ref}}}$ for $i$-th vehicle is defined using \eqref{x_ref}, \eqref{y_ref}, \eqref{psi_ref} and \eqref{v_ref}. Note that the generated reference trajectory is a naive initialization that might collide with other vehicles. The low-level planner (the proposed Distributed Algorithm \ref{alg_coordination}) ensures collision avoidance among the vehicles and at the same time satisfies all other types of physical constraints including dynamics, road boundaries, and actuation limits. \figref{fig:ref_generation} shows the generated reference trajectories for the reconfiguration scenario example \figref{fig:snapshot_4vehivle}. The reference trajectories of cyan and green vehicles are not straight lines, since they change their lanes. Note that the parameter $\rho$ determines when the lane change starts. For example, $\rho = 0.5$ means the lane change is performed in the middle of the total duration of the maneuver.

\subsubsection{Vehicle Shape} The $i$th vehicle dimensions are chosen as length $h = 4.5$m and width $w = 1.8$m. The road width is chosen as $3.7$m (i.e., the standard highway-lane width in the United States). The speed is lower bounded by zero. The acceleration of each vehicle is limited to a range of $\pm\,4$m/s${}^2$, and the rate of change of acceleration is bounded within $\pm\, 1$m/s${}^3$. The steering input is bounded within $\pm\,0.3$rad, and its rate change within $\pm 0.2$rad/s. 
The corresponding road region occupied by the $i$th vehicle is defined by a two-dimensional convex polytope $\mathcal{P}.$ For each vehicle, the transformed polytope is defined as $\mathcal{P}(\mathbf{z}(t)) = \{p
\in \mathbb{R}^{2}
|\mathbf{A}(\mathbf{z}(t))p\leq \mathbf{b}(\mathbf{z}(t))\},$ where 
$\mathbf{A}(\mathbf{z}(t))$ and $\mathbf{b}(\mathbf{z}(t))$ are defined as
$
\mathbf{A}(\mathbf{z}(t)) = 
\begin{bmatrix}
\mathbf{R}(\mathbf{z}(t))^{\top}\\
-\mathbf{R}(\mathbf{z}(t))^{\top}
\end{bmatrix},\
\mathbf{R}(\mathbf{z}(t)) = 
\begin{bmatrix} 
\cos(\psi(t))& -\sin(\psi(t))\\
\sin(\psi(t)) & \cos(\psi(t)) 
\end{bmatrix},
$ $\mathbf{b}(\mathbf{z}(t)) = [
h/2,w/2,h/2,w/2]^\top$ $+ \mathbf{A}(\mathbf{z}(t))[
x(t),
y(t)
]^{\top}$, where $[x(t),
y(t)
]^{\top}$ is the center of gravity of the vehicle.

\subsubsection{Formation Merging Results}
In this scenario, as seen in \figref{fig:snapshot_4vehivle}, the vehicles are initially traveling in three different lanes and need to merge in one lane to form a train-like platoon, while maintaining the safe distance $d_{min} = 0.5$m from each other at all times (i.e., during the lane change maneuvers and afterward). We tested the formation scenario for different configurations and initial conditions. \figref{fig:results_platoon_reconfiguration} shows an example with four vehicles. The initial longitudinal coordinates for all the four vehicles are $[x^1(0),x^2(0),x^3(0),x^4(0)]=[11.5,5.5,0.5,20]$ and the initial lateral coordinates are $[y^1(0),y^2(0),y^3(0),y^4(0)]=[1.85,5.55,1.85,9.25]$. The planning horizon $N$ is $0.75$s, the sampling time $\Delta t$ is $0.05$s, and $v_{\text{Ref}}$ is $15$m/s. \figref{fig:results_platoon_reconfiguration} represents the vehicles' states and actions. The longitudinal and lateral coordinates $x$ and $y$, as well as heading angle $\psi$ and velocity $v$ for all the vehicles are shown in different colors which are matched with the colors in \figref{fig:snapshot_4vehivle}. 

The control actions $a$ and $\delta$ are also illustrated for all the vehicles. The acceleration and velocity plots highlight the collaborative behavior between the vehicles. In contrast to reactive approaches, such as velocity obstacles, the speed of each vehicle is not assumed constant and varies based on the interactions with the neighbors. For example, as seen in the acceleration plot, the blue vehicle brakes and the magenta vehicle accelerates to make enough gap for other vehicles to merge into the lane. This behavior is obtained by solving the optimization problem and is not enforced explicitly. This collaborative behavior is fundamental to keeping the desired minimum distance and forming the platoon. Ellipsoidal representations of the vehicles would have required a larger minimum distance (to fit each vehicle the axes of each ellipse would have been 6.36m and 2.54m, respectively, leading to a minimum distance, in the longitudinal direction for example, equal to ($2(6.36-4.5)$m$>d_{\min}$) leading to more conservative behaviors. Similar considerations hold for implementations based on potential fields. A potential field can be always included in the NMPC problem formulation to enforce clearance with respect to the other vehicles, but it would lead to more conservative behaviors (e.g., larger minimum distance between the vehicles) and additional tuning parameters. In \figref{fig:snapshot_4vehivle}(a), in addition to snapshots, the separating hyperplanes between pink/green, green/red and red/blue pairs are shown and the normal of these separating hyperplanes are denoted as $\mathbf{s}_{pg}$, $\mathbf{s}_{gr}$ and $\mathbf{s}_{rb}$, respectively. 

\subsubsection{Performance Evaluation}
In this section, we provide various simulation analyses to evaluate the performance of the proposed algorithm. The two performance criteria that we study are computation efficiency and tight maneuvering since there is a trade-off between tight maneuvering and computation efficiency.
To evaluate the performance in terms of computation time of Algorithm~\ref{alg_coordination}, we simulated the formation of two, three, and four vehicles on a highway (varying the initial configuration of the vehicles to test the robustness of the tuning). We compared the average and maximum computation time in seconds for each step along the trajectory with the centralized implementation. Table \ref{tab:result_table1} shows the results.
\begin{table}
\begin{tabular}{|c?{0.5mm}c|c?{0.5mm}c|c|c|c|}
\hline
&\multicolumn{2}{c?{0.5mm}}{\textbf{Centralized}}& \multicolumn{4}{c|}{\textbf{Distributed}} \\
&\multicolumn{2}{c?{0.5mm}}{\textbf{}} & \multicolumn{2}{c|}{\textbf{NMPC}} &
\multicolumn{2}{c|}{\textbf{CA}} \\
\hline
\# & Avg.& Max.& Avg.& Max.& Avg.& Max.
\\
\hline
\multirow{3}{*}{2}& \multirow{3}{*}{1.3851} & \multirow{3}{*}{2.7482} &
0.1457 & 0.3621 & 0.0022 & 0.0024\\
&  &  & 0.1102 & 0.3313 & 0.0021 & 0.0022 \\\cline{4-7}
&  &  & \multicolumn{4}{c|}{\text{Total Avg. = 0.1301}}\\
\hline
\multirow{4}{*}{3}& \multirow{4}{*}{2.7680} & \multirow{4}{*}{4.6817} &
0.1848 & 0.3933 & 0.0025 & 0.0028\\
&  &  & 0.1206 & 0.3169 & 0.0024 & 0.0026 \\
&  &  & 0.1416 & 0.3470 & 0.0022 & 0.0023 \\\cline{4-7}
&  &  & \multicolumn{4}{c|}{\text{Total Avg. = 0.1514}}\\
\hline
\multirow{5}{*}{4}& \multirow{5}{*}{12.8331} & \multirow{5}{*}{28.7763} &
0.1919 & 0.4211 & 0.0030 & 0.0024\\
&  &  & 0.1125 & 0.2602 & 0.0027 & 0.0029\\
&  &  & 0.1842 & 0.3497 & 0.0024 & 0.0025  \\
&  &  & 0.2041 & 0.4195 & 0.0025 & 0.0023 \\\cline{4-7}
&  &  & \multicolumn{4}{c|}{\text{Total Avg. = 0.1757}}\\
\hline
\end{tabular}
\caption{Computation time (in second) for both centralized and distributed approaches are reported for two, three, and four numbers of coordinated vehicles.}
\label{tab:result_table1}
\end{table}
While both centralized and distributed approaches ensure safe vehicle coordination, the maximum and average computation time is much higher for the centralized approach compared to the distributed proposed algorithm. For example, the first row of Table \ref{tab:result_table1} shows a two-vehicle merging scenario. For this case, the average computation time for the centralized approach is $1.3851 (sec)$, while the average computation time for the proposed distributed algorithm is $0.1301 (sec)$. Therefore, even for the simplest case of a two-vehicle cooperative lane merging scenario, the centralized method's computation time renders it unsuitable for online real-world implementation. However, our proposed distributed algorithm exhibits an average computation time suitable for online implementation. By comparing the rest of the rows reported in Table \ref{tab:result_table1}, we can see the same behavior in computation time. The results are reported for three-vehicle cooperative merging as well as four-vehicle cooperative merging. Another criterion that can be studied in Table \ref{tab:result_table1} is scalability. Scalability in the context of centralized/distributed multi-agent schemes refers to the ability of a system to handle an increasing number of agents or participants efficiently without a significant degradation in performance. This criterion can be considered by looking at the columns of Table \ref{tab:result_table1}. As seen, for the centralized approach by increasing the number of agents from two to three and four, the average computation time increases dramatically, compromising the safety of the approach in real-time applications (i.e., delays in the calculation of the planning strategy can lead to collisions). However, looking at the columns of the distributed algorithm in Table \ref{tab:result_table1}, the average computation time is not affected significantly by adding more agents, and not only is the proposed distributed algorithm computationally efficient for online applications but also it is scalable to more number of agents. The distributed approach outperforms the centralized design by more than 2 orders of magnitude (when we look at the worst-case scenario with 4 vehicles) and the computation time is reasonable for online implementation.


\begin{figure}
 \vfill
\makebox[\columnwidth][c]{
\begin{subfigure}[a]{\columnwidth}
    \centering
   \includegraphics[width = \columnwidth]{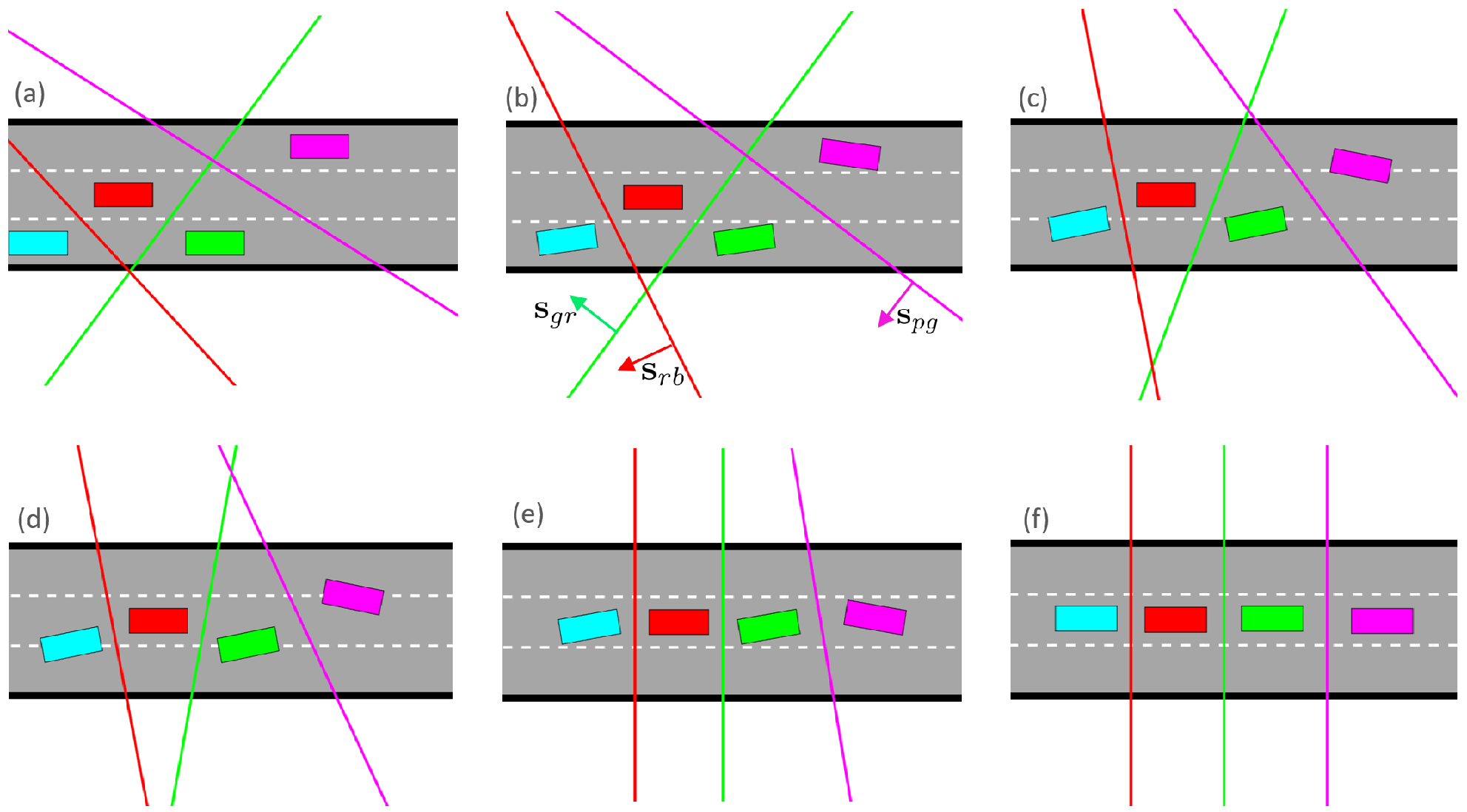}
   \caption{Snapshots}
   \label{fig:snapshot_4vehivle}
   \end{subfigure}
}\hfill\makebox[\columnwidth][c]{
    \begin{subfigure}[b]{\columnwidth}
    \includegraphics[width=\columnwidth]{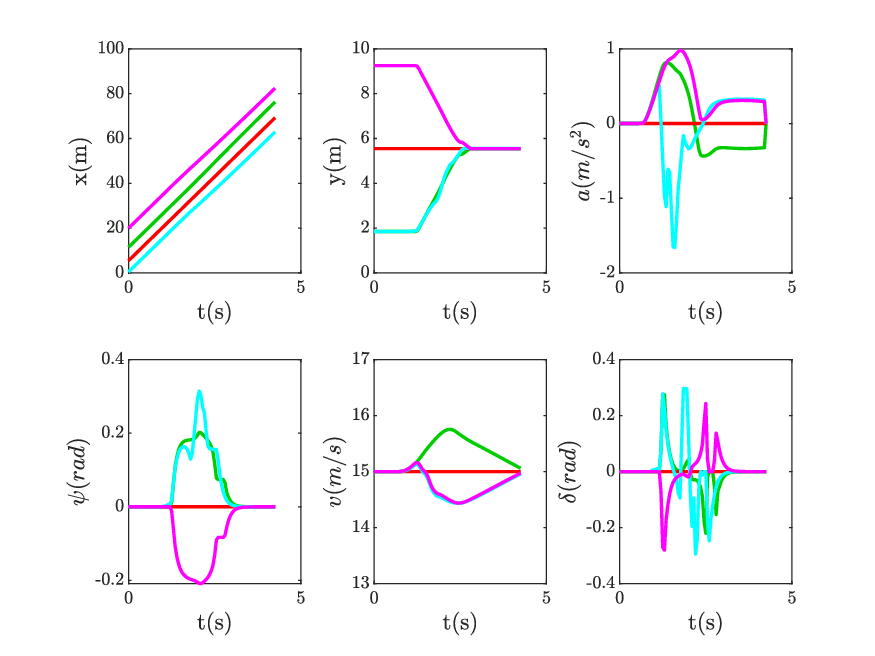}
   \caption{Vehicles trajectories}
   \label{fig:results_platoon_reconfiguration}
   \end{subfigure}
   }
   \caption{Four vehicles merge into a platoon in the center lane. The direction of motion is to the right. (a) Snapshots of the simulation are shown with the separating hyperplanes between the vehicles. (b) The state and input trajectories are shown.
}
\end{figure}

\begin{figure}
    \centering
    \includegraphics[width = \columnwidth]{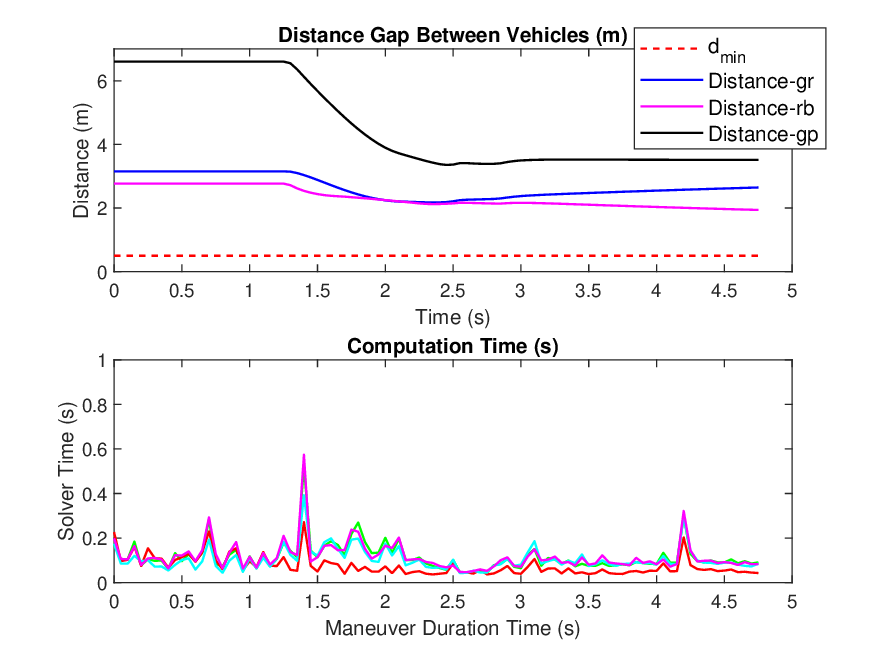}
    \caption{ Top: Distance between each pair of vehicles is shown during the cooperative maneuvers. (Distance-gr is the distance between green and red vehicles. Distance-rb is the distance between red and blue vehicles. Distance-gp is the distance between green and pick vehicles. $d_{\textrm{min}}$ is the minimum allowed distance.) Bottom: The solver computation time is reported for each vehicle along the entire maneuver duration.}
    \label{fig:gap_solver}
\end{figure}

For the same scenario of four-vehicle merging from multiple lanes (shown in \figref{fig:snapshot_4vehivle}), \figref{fig:gap_solver}, the top plot, shows the distance between three pairs of vehicles, which are in close proximity to each other. (The vehicles that are not reported are the ones with larger distances). Distance-gr denotes the distance between green and red vehicles, similarly, Distance-rb is the distance between red and blue vehicles, and Distance-gp represents the distance between green and pick vehicles. The bottom plot shows the solver computation time in seconds during the cooperative lane change maneuver of \figref{fig:snapshot_4vehivle}. As reported in the plot, the solver computation time increases when the vehicles are closest to each other. Since the solution space shrikes as the cars come closer to each other, the solver requires a longer time to obtain a solution.

{Although according to Table \ref{tab:result_table1}, the distributed approach outperforms the centralized approach in terms of computational efficiency and scalability, there is a trade-off between computation efficiency and how tight the maneuver can be. Solving the problem in a distributed way reduces the solution space by fixing the dual variables and solving for primal ones and therefore results in less tight trajectories for vehicles compared to the centralized approach. In Table \ref{tab:result_table2} the total cost of the entire maneuver is reported for centralized and distributed approaches for two, three, and four numbers of coordinated vehicles. The cost is the summation of the closed-loop costs for the entire maneuver. The cost is calculated based on the cost defined in \eqref{eq:cost}. As seen, the sum of costs for the distributed approach is larger compared to the one for the centralized approach. Although the total cost for the distributed approach is higher compared to the centralized one, still the gaps between vehicles are sufficiently small and the resulting maneuver is tight such that other non-cooperative vehicles in the surrounding traffic cannot merge into these cooperative cars.  
}

\begin{table}
\centering
\begin{tabular}{|c?{0.5mm}c?{0.5mm}c|}
\hline
& {\textbf{Centralized}}& {\textbf{Distributed}} \\
\hline
\# & Sum of Costs & Sum of Costs\\
\hline
\multirow{3}{*}{2}& \multirow{3}{*}{0.0443} & 1.8957 \\
&  &  1.6845  \\ \cline{3-3}
&  &  \text{Total Avg. = 1.7901}\\
\hline
\multirow{4}{*}{3}& \multirow{4}{*}{0.0985} & 6.0267 \\
&  &  0.0000 \\
&  &  5.6856  \\\cline{3-3}
&  &  {\text{Total Avg. = 3.9041}}\\
\hline
\multirow{5}{*}{4}& \multirow{5}{*}{0.0321}  &
6.8064 \\
&  &   0.0000 \\
&  &   12.1444   \\
&  &   10.3252  \\\cline{3-3}
&  &  {\text{Total Avg. = 7.3190}}\\
\hline
\end{tabular}
\caption{The total cost (sum of the closed-loop costs for the entire maneuver) is reported for centralized and distributed approaches for two, three, and four numbers of coordinated vehicles. }
\label{tab:result_table2}
\end{table}

\begin{figure}
    \centering
    \includegraphics[width = \columnwidth]{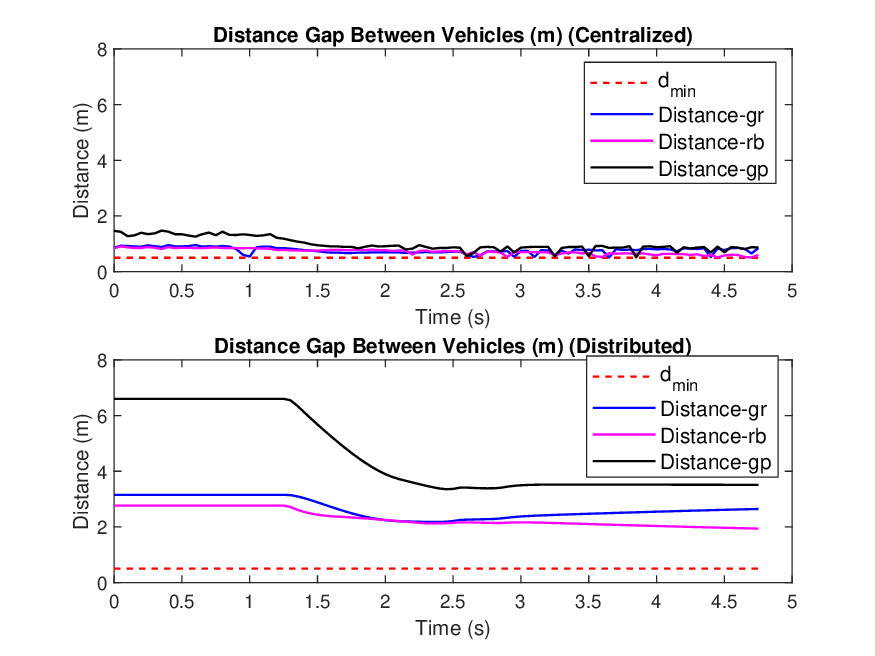}
    \caption{ Top: The distance between each pair of vehicles is shown during the cooperative maneuvers for centralized formulation. Bottom: The distance between each pair of vehicles is shown along the entire maneuver for distributed formulation. (Distance-gr is the distance between green and red vehicles. Distance-rb is the distance between red and blue vehicles. Distance-gp is the distance between green and pick vehicles. $d_{\textrm{min}}$ is the minimum allowed distance.)}
    \label{fig:distance_gap_compare}
\end{figure}

\begin{figure}
    \centering
    \includegraphics[width = \columnwidth]{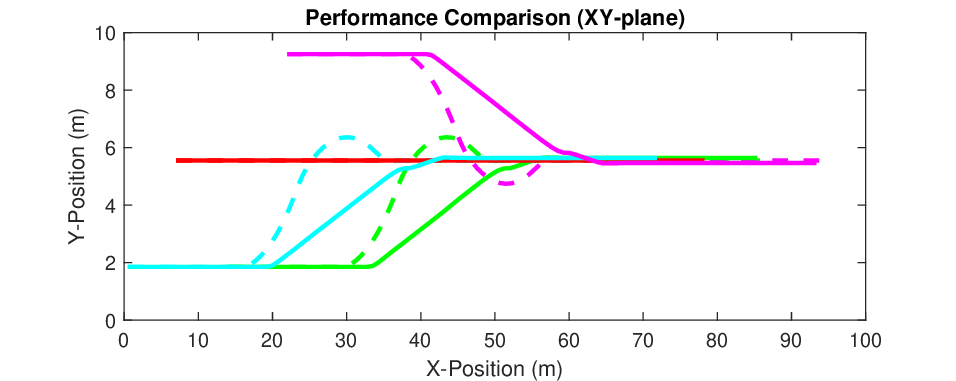}
    \caption{The trajectories in XY plane are shown for both centralized (in dashed lines) and distributed (solid lines) to visualize the same simulation of cooperative lane-change from multiple lanes shown in \figref{fig:snapshot_4vehivle} to compare the maneuver tightness in centralized and distributed algorithms. As seen centralized cooperative maneuver is tighter compared to the distributed algorithm since the dashed lines converge faster. The colors correspond to the same simulation of cars with the same colors in \figref{fig:snapshot_4vehivle}. }
    \label{fig:xy_compare}
\end{figure}

In \figref{fig:distance_gap_compare} and \figref{fig:xy_compare}, the performance of centralized and distributed approaches are compared in terms of the tightness of maneuver. In \figref{fig:distance_gap_compare}, the performance of centralized and distributed methods are compared based on the distance between the cars during the cooperative multi-lane change maneuver. The top plot shows the distance between pairs of cars. As shown these distances are small and they all satisfy the minimum distance $d_{\textrm{min}}$ which is $0.5 (m)$ in this simulation. In the bottom plot, the pair-wise distances among cars are reported for the distributed algorithm. As seen, these distances also respect the minimum distance $d_{\textrm{min}}$ over the entire maneuver, but they are larger compared to the centralized method. The distributed approach cannot achieve maneuvers as tightly as the centralized approach. In \figref{fig:xy_compare}, the entire maneuver is shown in the XY plane. As seen, the centralized maneuvers (in dashed lines) are tight compared to the maneuvers obtained from the distributed algorithm. This is the limitation or disadvantage of the proposed distributed algorithm that due to the alternation scheme cannot provide as tight maneuvers as the centralized framework. However, the distances provided by the distributed method are still suitable for the tight formation of cooperative cars in online applications.

\section{CONCLUSIONS}
\label{SEC7}

We proposed a distributed algorithm for multi-vehicle coordination in tight spaces using nonlinear MPC and strong duality theory. We reformulated the collision avoidance constraints in the dual formulation and used dual decomposition to split the large centralized optimization problem into smaller sub-problems.
We showed the effectiveness of the algorithm for the coordination of connected and automated vehicles on public roads through platoon merging. Our results show that the distributed algorithm is computationally efficient for online implementation and is scalable to larger networks of vehicles. According to our simulation studies, the distributed approach outperforms the centralized design by more than 2 orders of magnitude. Moreover, the outcomes indicate a trade-off between the level of tightness achievable in a maneuver and the computation time required for optimization. While the distributed approach may result in maneuvers that are not as tight as those in the centralized approach, it still offers proximity maneuvers that are well-suited for online implementation. 

As part of future work, we plan to extend this work to 3-dimensional planning, since the proposed method can be used to avoid collisions in $3$-dimensional spaces. We will consider uncertainty due to communication delay, model mismatch, and sensor measurements Also, we plan to test with real hardware (e.g., drones and ground vehicles) and in more complex scenarios (e.g., urban driving). From the algorithm-design perspective, we plan to investigate strategies to deal with real sensor data, communication delays, and random faults.

\section{APPENDIX}
\label{SEC8}

\subsection{KKT Conditions for Geometric Interpretations:} The geometric interpretation of dual variables can be explained by obtaining the Karush-Kuhn-Tucker (KKT) conditions for problem \eqref{eq:pprimal}.
The dual problem expressed in \eqref{eq:dual}
\begin{align}
    &\mathbf{A}_1 = 
    \begin{bmatrix} 
    \horzbar \mathbf{a_1}_{1} \horzbar \\ 
    \horzbar \mathbf{a_1}_{2} \horzbar \\
    \vdots\\
    \horzbar \mathbf{a_1}_{m_1} \horzbar
    \end{bmatrix}, \quad
    \mathbf{A}_2 = 
    \begin{bmatrix} 
    \horzbar \mathbf{a_2}_{1} \horzbar \\ 
    \horzbar \mathbf{a_2}_{2} \horzbar \\
    \vdots\\
    \horzbar \mathbf{a_2}_{m_2} \horzbar
    \end{bmatrix}, 
\end{align}
\begin{align}
    &\mathbf{b}_1^\top = 
    \begin{bmatrix} 
    b_{11} &
    \hdots &
    b_{1m_1}
    \end{bmatrix}, 
    \mathbf{b}_2^\top = 
    \begin{bmatrix} 
    b_{21} &
    \hdots &
    b_{2m_2}
    \end{bmatrix},
\end{align}
\begin{align}
    &\boldsymbol{\lambda}_{12}^\top = 
    \begin{bmatrix} 
    \lambda_{121} &
    \lambda_{122} &
    \hdots &
    \lambda_{12m_1}
    \end{bmatrix}, \\
    &\boldsymbol{\lambda}_{21}^\top = 
    \begin{bmatrix} 
    \lambda_{211} &
    \lambda_{212} &
    \hdots &
    \lambda_{21m_2}
    \end{bmatrix}.
\end{align}
Since the primal problem is a convex function, the KKT condition is necessary and sufficient for the points to be primal and dual optimal. Other than the feasibility of primal and dual problems, there is complementary slackness
\begin{align}
    &\lambda_{12n}(\mathbf{a_1}_{n}\mathbf{x} - b_{1n}) = 0, \quad n\in\{1,2,\hdots,m_1\}, \label{cs1}\\
    &\lambda_{21n}(\mathbf{a_2}_{n}\mathbf{y} - b_{2n}) = 0, \quad n\in\{1,2,\hdots,m_2\}, \label{cs2}
\end{align}
By expanding \eqref{cs1} and \eqref{cs2} we will have
\begin{align}
    &\mathbf{x}^\top \mathbf{a_1}_{n}^\top \lambda_{12n} = b_{1n} \lambda_{12n},\ n\in\{1,2,\hdots, m_1\} \label{eq1},\\
    &\mathbf{y}^\top \mathbf{a_2}_{n}^\top \lambda_{21n} = b_{2n} \lambda_{21n},\ n\in\{1,2,\hdots, m_2\} \label{eq2}
\end{align}
where $\mathbf{a_1}_{n}$ and $\mathbf{a_2}_{n}$ are row vectors of $\mathbf{A}_1$ and $\mathbf{A}_2$, respectively. The other KKT condition is the stationarity condition which is based on the primal problem in \eqref{eq:pprimal}
\begin{align}
    &\mathbf{A}_1^\top\boldsymbol{\lambda}_{12}+\mathbf{s} = 0, \label{sc1}\\
    &\mathbf{A}_2^\top\boldsymbol{\lambda}_{21}-\mathbf{s} = 0, \label{sc2} \\
    &\frac{\mathbf{w}}{\|\mathbf{w}\|}-\mathbf{s} = 0. \label{sc3}
\end{align}
As seen, \eqref{sc1} and \eqref{sc2} reflect the equality constraints of the dual problem \eqref{eq:dual}. Combining all together, at optimum we have $\mathbf{s}^\top\mathbf{x} = -\mathbf{b}_1^\top\boldsymbol{\lambda}_{12} > 0$ and $\mathbf{s}^\top\mathbf{y} = +\mathbf{b}_2^\top\boldsymbol{\lambda}_{21} < 0$.

\bibliographystyle{IEEEtran}

\bibliography{bibliography}
\end{document}